# Brain Cancer Segmentation Using YOLOv5 Deep Neural Network

Version 1

Sudipto Paul
ID: 201-15-3478
Department of Computer Science & Engineering
Daffodil International University
Dhaka, Bangladesh
sudipto15-3478@diu.edu.bd

Dr. Md Taimur Ahad
*Associate Professor*
Department of Computer Science & Engineering
Daffodil International University
Dhaka, Bangladesh
taimur.cse0396@diu.edu.bd

Md. Mahedi Hasan
*Researcher*
Department of Computer Science & Engineering Daffodil International University
Dhaka, Bangladesh
mh.cse9@gmail.com

*Abstract* — An expansion of aberrant brain cells is referred to as a brain tumor. The brain's architecture is extremely intricate, with several regions controlling various nervous system processes. Any portion of the brain or skull can develop a brain tumor, including the brain's protective coating, the base of the skull, the brainstem, the sinuses, the nasal cavity, and many other places. Over the past ten years, numerous developments in the field of computer-aided brain tumor diagnosis have been made. Recently, instance segmentation has attracted a lot of interest in numerous computer vision applications. It seeks to assign various IDs to various scene objects, even if they are members of the same class. Typically, a two-stage pipeline is used to perform instance segmentation. This study shows brain cancer segmentation using YOLOv5. Yolo takes dataset as picture format and corresponding text file. You Only Look Once (YOLO) is a viral and widely used algorithm. YOLO is famous for its object recognition properties. You Only Look Once (YOLO) is a popular algorithm that has gone viral. YOLO is well known for its ability to identify objects. YOLO V2, V3, V4, and V5 are some of the YOLO latest versions that experts have published in recent years. Early brain tumor detection is one of the most important jobs that neurologists and radiologists have. However, it can be difficult and error-prone to manually identify and segment brain tumors from Magnetic Resonance Imaging (MRI) data. For making an early diagnosis of the condition, an automated brain tumor detection system is necessary. The model of the research paper has three classes. They are respectively Meningioma, Pituitary, Glioma. The results show that, our model achieves competitive accuracy, in terms of runtime usage of M2 10 core GPU.

## I. INTRODUCTION

YOLO is a method that provides real-time object detection using neural networks. The popularity of this algorithm is due to its accuracy and quickness. It has been applied in a variety of ways to identify animals, humans, parking meters, and traffic lights. The following justifications make the YOLO algorithm crucial:

- Speed: Because this algorithm can predict objects in real-time, it increases the speed of detection.

- High accuracy: The YOLO prediction method yields precise findings with few background mistakes.

- Learning capabilities: The algorithm has great learning capabilities that allow it to pick up on object representations and use them for object detection.

Yolo works using the three techniques; Residual blocks, Bounding box regression, Intersection Over Union (IOU). Early identification and treatment planning are essential for a correct diagnosis of a brain tumor. Medical image analysis requires careful consideration of digital image processing. Separating aberrant brain tissues from normal brain tissues is the process of segmenting a brain tumor. The early detection of brain cancer reduces the impact of surgery and treatment, improving the prognosis for many patients. Brain tumors can affect your brain function whether or not they are cancerous if they enlarge to the point where they press against nearby tissues, so, it is much important to early detect brain cancer.

The task of segmenting and detecting brain tumors has been heavily automated, either partially or completely, by a number of studies:

A study was done on a subset of the BRATS 2018 dataset that contained 1,992 Brain MRI scans. The YOLOv5 model achieved an accuracy of 85.95% and the Fast Ai classification model achieved an accuracy of 95.78%. These two models can be applied in real-time brain tumor detection for early diagnosis of brain cancer [1].

Another work was presented is prepared with the 29-layer YOLO Tiny and fine-tuned to work efficiently and perform task productively and accurately in most cases with solid execution. The outcome of the model is the highest like precision, recall and F1-score beating other previous results of earlier versions of YOLO and other studies like Fast R-CNN [2].

## II. LITERATURE REVIEW

**YOLOv5**

The You Only Look Once (YOLO) family of computer vision models includes the model known as YOLOv5. YOLOv5 is frequently employed for object detection. Small (s), medium (m), large (l), and extra-large (x) are the four major variants of YOLOv5, each of which offers increasingly higher accuracy rates. A variable amount of time is required to train for each type. The YOLOv5 repository is a logical progression from Glenn Jocher's YOLOv3 PyTorch repository. Developers frequently ported the YOLOv3 Darknet weights to PyTorch in the YOLOv3 PyTorch repository before proceeding to production. Many people loved how simple the PyTorch branch was to use and would use this outlet for deployment. With each training batch, YOLOv5 passes training data through a data loader, which augments data online. The data loader makes three kinds of augmentations:
- Scaling,
- Color space adjustment,
- Mosaic augmentation.

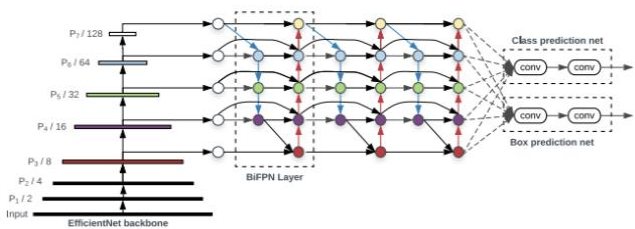

Figure 1: Architecture and object detection of YOLO

**Brain Cancer**

A lump or growth of abnormal cells in your brain is known as a brain tumor. There are numerous varieties of brain tumors. Both benign (noncancerous) and malignant (cancerous) brain tumors can occur (malignant). Primary brain tumors are those that start in the brain; secondary (metastatic) brain tumors are those that start in other regions of the body and spread to the brain. Primary brain tumors come in many distinct varieties. Each is named for the class of cells that are involved. Examples comprise:
- Gliomas: These tumors begin in the brain or spinal cord and include astrocytomas, ependymomas, glioblastomas, oligoastrocytomas and oligodendrogliomas.
- Meningiomas: A meningioma is a tumor that arises from the membranes that surround your brain and spinal cord (meninges). Most meningiomas are noncancerous.
- Acoustic neuromas (schwannomas): These are benign tumors that develop on the nerves that control balance and hearing leading from your inner ear to your brain.
- Pituitary adenomas: These are tumors that develop in the pituitary gland at the base of the brain. These tumors can affect the pituitary hormones with effects throughout the body.
- Medulloblastomas: These cancerous brain tumors are most common in children, though they can occur at any age. A medulloblastoma starts in the lower back part of the brain and tends to spread through the spinal fluid.
- Germ cell tumors: Germ cell tumors may develop during childhood where the testicles or ovaries will form. But sometimes germ cell tumors affect other parts of the body, such as the brain.
- Craniopharyngiomas: These rare tumors start near the brain's pituitary gland, which secretes hormones that control many body functions. As the craniopharyngioma slowly grows, it can affect the pituitary gland and other structures near the brain.

Secondary (metastatic) brain tumors are tumors that develop when cancer spreads (metastasizes) to the brain from another part of the body. People with a prior history of cancer are more likely to develop secondary brain tumors. Rarely, a brain tumor that has spread to other parts of your body may be the first sign of cancer.

In this section we review some models for object detection and reviewing:

Classification and object identification is the main objective of this research work. The darknet yolov4 is used, to perform the classification, and region of interest detection with the best accuracy scores. The model is trained with the Tesla GPU and obtained the results of the existing techniques in the field of fetal brain classification and localization. The accuracy of 97.92% and precision percentage of 96.70 is achieved in the research work [3].

The models were trained on 641 MRI scan images taken from this dataset. After evaluating the experimental results of these models, we determined that the YOLO V5 model provided the best performance as it was able to reach a mAP@0.5 score of 95.07%. In contrast, the YOLO V3 Pytorch model provided the worst accuracy as it earned a mAP@0.5 score of 84.30%. Real-time implementations of these models can provide medical professionals with a highly efficient, automatic brain tumor diagnostic tool that will revolutionize the field of neuroscience [1].

In this paper, the Precise Epic Localization Algorithm was adopted in YOLO v4 architecture to detect and classify the healthy fetal brain with its orientation and unhealthy brain with their abnormalities from the given input of MRI Images. The detection and classification of Encephalocele and Arteriovenous Malformation from a fetal brain MRI are obtained and evaluated using a machine learning algorithm to determine the abnormalities with the accuracy of 97.27%, which outperforms the public tools, BET and ROBEX. As Tesla P100 GPU is employed in the cloud environment, the output is more convenient and accessible than the existing methods [4].

This paper presents a deep learning-based approach for brain tumor identification and classification using the state-of-the-art object detection framework YOLO (You Only Look Once). The YOLOv5 is a novel object detection deep learning technique that requires limited computational architecture than its competing models. The study used the Brats 2021 dataset from the RSNA-MICCAI brain tumor radio genomic classification. The dataset has images annotated from RSNA-MICCAI brain tumor radio genomic competition dataset using the make sense an AI online tool for labeling dataset. The preprocessed data is then divided into testing and training for the model. The YOLOv5 model provides a precision of 88 percent. Finally, the model is tested across the whole dataset, and it is concluded that it is able to detect brain tumors successfully [5].

This work trained with the partial 29-layer YOLOv4-Tiny and fine-tuned to work optimally and run efficiently in most platforms with reliable performance. With the help of transfer learning, the model had initial leverage to train faster with pre-trained weights from the COCO dataset, generating a robust set of features required for brain tumor detection. The results yielded the highest mean average precision of 93.14%, a 90.34% precision, 88.58% recall, and 89.45% F1-Score outperforming other previous versions of the YOLO detection models and other studies that used bounding box detections for the same task as Faster R-CNN. As concluded, the YOLOv4-Tiny can work efficiently to detect brain tumors automatically at a rapid phase with the help of proper fine-tuning and transfer learning. This work contributes mainly to assist medical experts in the diagnostic process of brain tumors [6].

In this work, a novel Shuffled-YOLO network has been proposed for segmenting brain tumors from multimodal MRI images. Initially, the scalable range-based adaptive bilateral filer (SCRAB) pre-processing technique was used to eliminate the noise artifacts from MRI while preserving the edges. In the segmentation phase, we propose a novel deep Shuffled-YOLO architecture for segmenting the internal tumor structures that include non-enhancing, edema, necrosis, and enhancing tumors from the multi-modality MRI sequences. The experimental fallouts reveal that the proposed Shuffled-YOLO network achieves a better accuracy range of 98.07% for BraTS 2020 and 97.04% for BraTS 2019 with very minimal computational complexity compared to the state-of-the-art models [7].

In this paper, The results in the first stage show that the proposed regional-based YOLO efficiently detected the CMBs with an overall sensitivity of 93.62% and an average number of false positives per subject (FP avg ) of 52.18 throughout the five-folds cross-validation. The 3D-CNN based second stage further improved the detection performance by reducing the FP avg to 1.42. The outcomes of this work might provide useful guidelines towards applying deep learning algorithms for automatic CMBs detection [8].

Implementation of YOLOv2 and YOLOv3 is done on the own created data set for the real-time UAV's detection and to benchmark the performance of both models in terms of mean average precision (MAP) and accuracy. For the specifically created data set made, YOLOv3 is outperforming YOLOv2 both in MAP and accuracy [9].

This paper shows us, the evaluation results of the YOLO-based detection achieved detection accuracy of 97.27%, Matthews's correlation coefficient (MCC) of 93.93%, and F1-score of 98.02%. Moreover, the results of the breast lesion segmentation via FrCN achieved an overall accuracy of 92.97%, MCC of 85.93%, Dice (F1-score) of 92.69%, and Jaccard similarity coefficient of 86.37%. The detected and segmented breast lesions are classified via CNN, ResNet-50, and InceptionResNet-V2 achieving an average overall accuracy of 88.74%, 92.56%, and 95.32%, respectively. The performance evaluation results through all stages of detection, segmentation, and classification show that the integrated CAD system outperforms the latest conventional deep learning methodologies [10].

This study is all about YOLO and convolution neural networks (CNN)in the direction of real time object detection.YOLO does generalize object representation more effectively without precision losses than other object detection models.CNN architecture models have the ability to eliminate highlights and identify objects in any given image. When implemented appropriately, CNN models can address issues like deformity diagnosis, creating educational or instructive application, etc. This article reached at number of observations and perspective findings through the analysis. Also, it provides support for the focused visual information and feature extraction in the financial and other industries, highlights the method of target detection and feature selection, and briefly describe the development process of YOLO algorithm [11].

III. EXPERIMENT DESCRIPTION

In this research, we used YOLOv5 which is newer version of YOLO to segment. Our model provides us competitive results. Basically, YOLO takes dataset as picture and with it's corresponding text file. There is different folder for train and validation. (.yaml) extension file was taken to define path and classes. It's mandatory in newer YOLO versions. After following proper processing and segmentation we get results.

IV. RESELT OF EXPERIMENT 1: TRAIN AND VALIDATION RATIO

Train and validation ratio was taken 80% and 20%

```
training images are  :   720
Validation images are:   180
```

Figure: Output of quantity of train images and validation images

## V. RESELT OF EXPERIMENT 2: LABELS

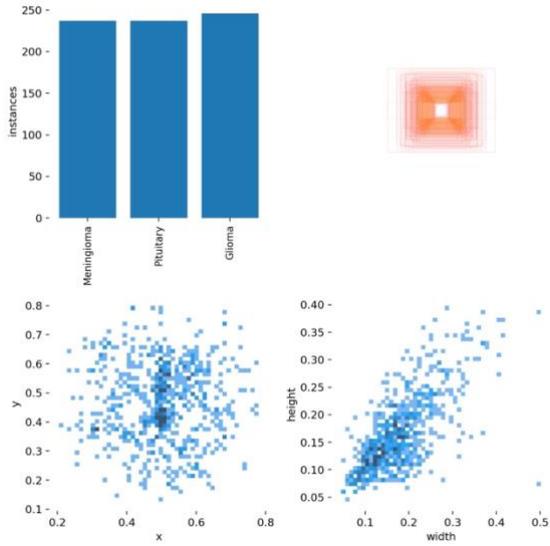

Figure: Labels

## VI. RESULT OF EXPERIMENT 3: LABELS CORRELOGRAM

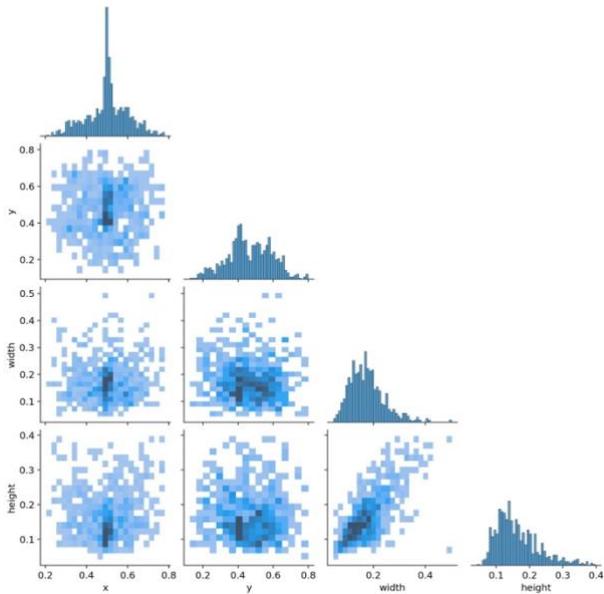

Figure: Labels Correlogram

## VII. RESULT OF EXPERIMENT 4: PR_CURVE

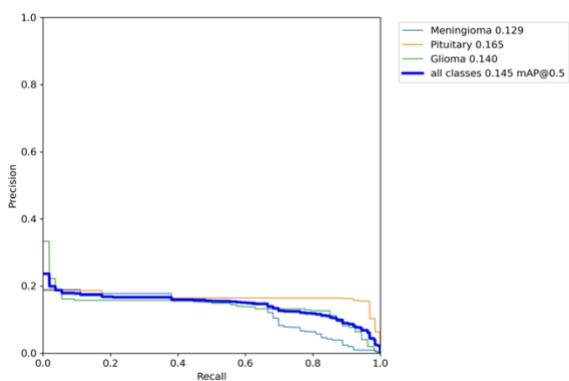

Figure: Precision recall curve

## VIII. RESULT OF EXPERIMENT 5: TRAIN_BATCH1

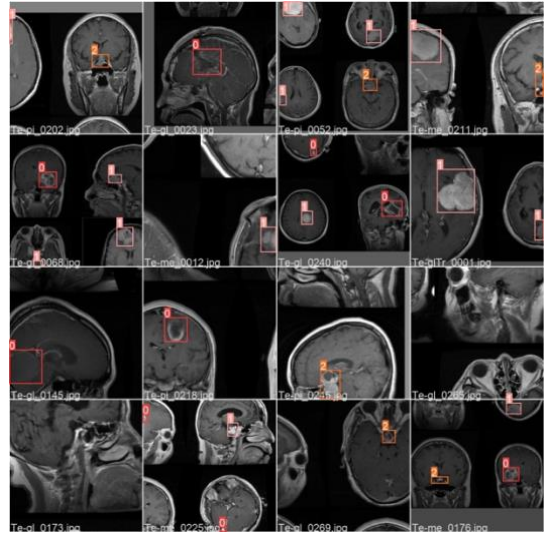

Figure: Train batch 1 image

## IX. RESULT OF EXPERIMENT 6: TRAIN_BATCH2

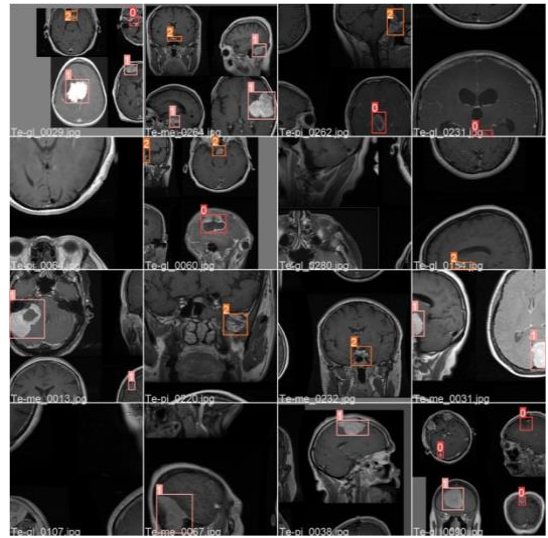

Figure: Train batch 2 image

## X. RESULT OF EXPERIMENT 7: VALIDATION BATCH LABELS

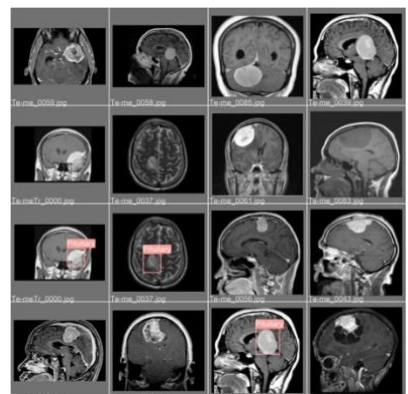

Figure: Image of validation batch labels

## XI. RESULT OF EXPERIMENT 8: VALIDATION BATCH PREDICTION

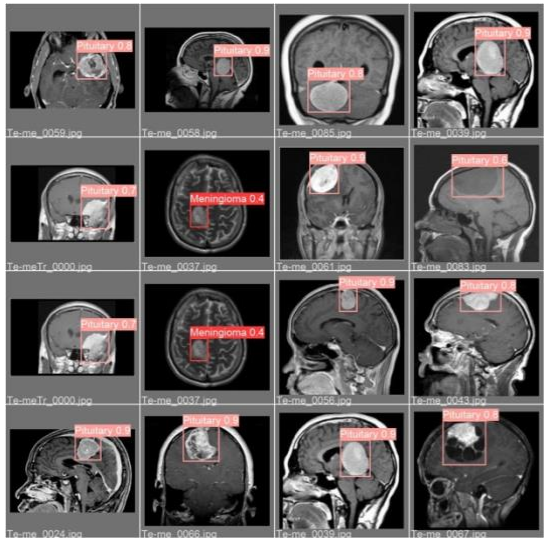

Figure: Image of validation batch prediction

## XII. RESULT OF EXPERIMENT 9: RESULT TABLE

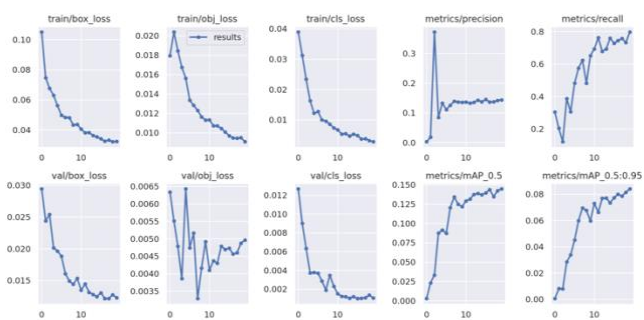

Figure: Result table

## XIII. RESULTS OF EXPERIMENT 10. RESULT IMAGE

Figure: Image of result

## XIV. RESULT OF EXPERIMENT 11. CONFUSION MATRIX

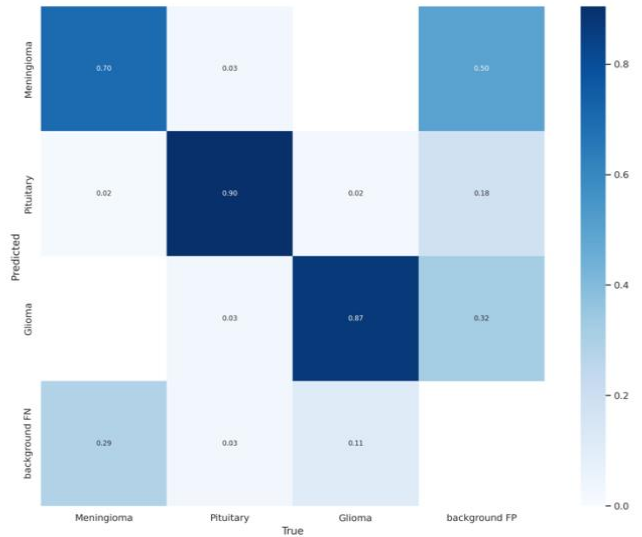

Figure: Confusion Matrix

## XV. CONTRIBUTIONS

A segmentation study of brain tumor is important to gain a full understanding of YOLO performance in brain tumor research using three classes' images, we presented segmentation. We used YOLOv5 for segmentation which is faster then previous versions.

## XVI. LIMITATIONS

In this study, a segmentation model was developed to accurately detect.

There are a number of limitations in the current stage of the research, which need to address in future work. The use of free-of-charge resources (Google Colab) limits the experiments of this study. As Google Colab offers the server for a limited time, the hyperparameter tuning, training the base model training other than Imagenet (this research used Imagenet as the base database in transfer learning), and the application of Adadelta, FTRL, NAdam, Adadelta, and many more optimizers were not performed in this study. Another limitation is that the research used secondary data that are publicly available, not primary data directly collected from fields.

In the future, we want to create a user interface for the detection and localization of rice leaf diseases for farmers. This interface would have not only detection but also provide a guide on how the diseases can be controlled. As mobile phones are seen as the preferred technological device among users in developing countries, our aim is to develop a mobile phone-based rice leaf disease detection application tool.

## XVII. CONCLUSION

For proper treatment and taking action, early detection of brain cancer is important to reduce the risk of brain cancer. Using YOLOv5 we created segmentation and got proper output. We have three classes and have proper segmentation output. From the output we can visualize brain cancer of accurate classes.

## XVIII. DECLARATION OF COMPETING INTEREST

The authors declare that they have no known competing financial interests or personal relationships that could have appeared to influence the work reported in this paper.

## XIX. FUNDING STATEMENT

The work was not supported by any funding and neither did any of the researchers receive funds.